\pdfoutput=1

\documentclass[11pt]{article}

\usepackage[final]{acl}

\usepackage{times}
\usepackage{latexsym}

\usepackage[T1]{fontenc}

\usepackage[utf8]{inputenc}

\usepackage{microtype}

\usepackage{hyperref}
\usepackage{url}
\usepackage{booktabs}
\usepackage{subcaption}
\usepackage{amsmath}
\usepackage{pgfplots}
\usepackage{multirow}
\usepackage{makecell}
\usepackage{wrapfig}
\usepackage{tcolorbox} 
\usepackage{xcolor} 
\usepackage{fvextra} 
\DefineVerbatimEnvironment{Verbatim}{Verbatim}{breaklines=true}

\usepackage{inconsolata}

\usepackage{graphicx}

\title{Is the Top Still Spinning? Evaluating Subjectivity\\in Narrative Understanding}

\author{
 \textbf{Melanie Subbiah\textsuperscript{1}},
 \textbf{Akankshya Mishra\textsuperscript{1}},
 \textbf{Grace Kim\textsuperscript{2}},
 \textbf{Liyan Tang\textsuperscript{2}},
\\
 \textbf{Greg Durrett\textsuperscript{2}},
 \textbf{Kathleen McKeown\textsuperscript{1}}
\\
\\
 \textsuperscript{1}Columbia University,
 \textsuperscript{2}The University of Texas at Austin
\\
 \small{
   \textbf{Correspondence:} \href{mailto:m.subbiah@columbia.edu}{m.subbiah@columbia.edu}
 }
}

\definecolor{myblue}{RGB}{0, 70, 140}

\begin{document}
\maketitle
\begin{abstract}
Determining faithfulness of a claim to a source document is an important problem across many domains. This task is generally treated as a binary judgment of whether the claim is supported or unsupported in relation to the source. In many cases, though, whether a claim is supported can be ambiguous. For instance, it may depend on making inferences from given evidence, and different people can reasonably interpret the claim as either supported or unsupported based on their agreement with those inferences. Forcing binary labels upon such claims lowers the reliability of evaluation. In this work, we reframe the task to manage the subjectivity involved with factuality judgments of ambiguous claims. We introduce LLM-generated edits of summaries as a method of providing a nuanced evaluation of claims: how much does a summary need to be edited to be unambiguous? Whether a claim gets rewritten and how much it changes 
can be used as an automatic evaluation metric, the Ambiguity Rewrite Metric (ARM), with a much richer feedback signal than a binary judgment of faithfulness. 
We focus on the area of narrative summarization as it is particularly rife with ambiguity and subjective interpretation. We show that ARM 
 produces a 21\% absolute improvement in annotator agreement on claim faithfulness
 , indicating that subjectivity is reduced. 
\end{abstract}

\section{Introduction}

A possible solution to the problem of factual errors in LLM-generated output lies in having a separate model or process to verify factuality \cite{durmus2020feqa, laban2022summac, chen2023felm, tang-etal-2024-minicheck}. In domains such as mathematical reasoning, verifiers like this can be used not only for evaluation, but at either training time or inference time to improve models \citep{zelikman2022star, wang2023math}. However, whether an output is factual, or whether it is entailed given some input, has been shown to be highly subjective \citep{pavlick2019inherent, nie2020can, jiang-marneffe-2022-investigating}.
This kind of subjectivity is common in tasks in the social sciences and humanities. In this work, we address narrative summarization, which plays a dual role of being a useful application of LLMs in and of itself as well as a proxy task for dealing with complex issues of subjectivity in factual judgments.


Summarizing a story is a method of capturing and distilling the key details and takeways from that narrative \citep{kryscinski2021booksum, chang2024booookscore}. In this way, summarization is a vehicle for examining the understanding the summarizer has of the story. Some understandings can be clearly wrong. For example, a summary of \textit{Pride and Prejudice} is wrong if it says that Mr. Darcy is a poor farmer.\footnote{He is a very wealthy member of the landed gentry.} 
Other understandings are matters of interpretation. For example, some might summarize the end of the movie \textit{Inception} as saying the top wobbles and is about to fall, indicating the main character has returned to reality,\footnote{The main character spins a top to check if he is in reality or another person's subconscious. If it falls, he is in the real world, whereas it will just keep spinning in the subconscious. At the end, it seems he is in reality but when he spins the top, the credits roll before we see if it falls.} while others would disagree with this interpretation. 

\begin{figure*}[t]
\centering
    \includegraphics[width=.8\linewidth]{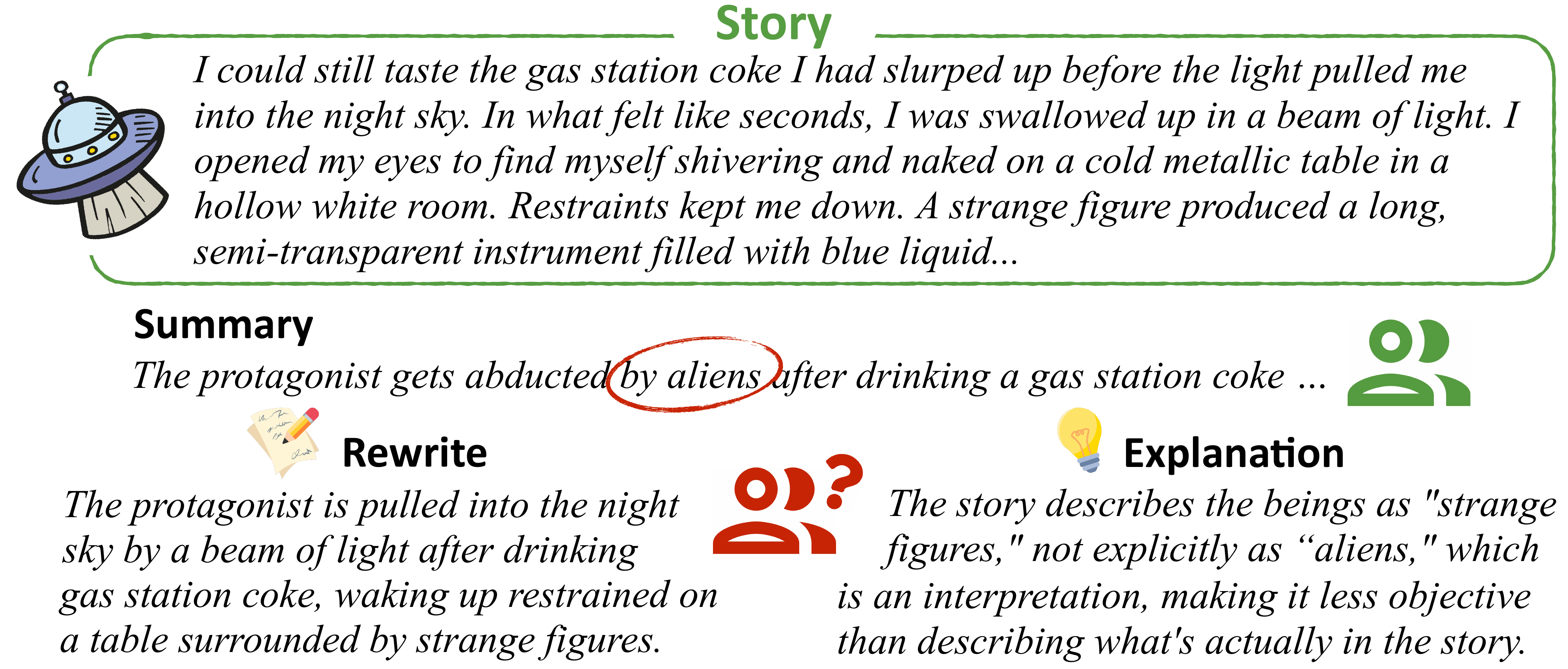}
    \caption{An example from StorySumm where the summary makes the interpretive leap that this story is describing an alien abduction. Many people find this to be a reasonable assumption and agree with it. Others correctly point out that the story never explicitly states these are aliens, only "strange figure(s)". The rewrite is more objectively faithful to the story. 
    }
    \label{fig:introexample}
\end{figure*}

Summarization work has traditionally evaluated summary faithfulness as a binary judgment of whether or not each detail in a summary is faithful to the source \citep{durmus2020feqa, fabbri-etal-2021-summeval, min2023factscore}. 
If the claim says the top is about to fall though, there will be disagreement on this binary label. With narratives, sometimes claim wording is ambiguous or interpretive in a way that leads to subjective judgments of faithfulness (see example in Figure \ref{fig:introexample}).
Prior work has shown that in practice, it is challenging for humans to agree on this binary judgment, let alone produce a reliable automatic evaluator for narrative summarization \citep{subbiah2024storysumm}. 

To resolve this, we first remove the assumption that there can be a universally agreed upon faithful/unfaithful label for all claims. 
We instead use LLM-generated rewrites as a method of evaluating claims, which we term the Ambiguity Rewrite Metric (ARM). 
This method produces a binary judgment (whether or not a claim is rewritten), as well as a correction of the issues in the claim. The degree to which the claim changed also provides signal on how flawed the original claim was. For example, instead of labeling the claim about the spinning top as true or false, a rewrite could  specify that the top wobbles and we do not see whether or not it falls. We can quantify the amount of rewriting either through edit distance or the number of explanation points necessary to justify the changes. 


We additionally uncover the types of ambiguities in narrative claims that lead to disagreements. We distinguish between intentional ambiguity introduced by the story author and unintentional ambiguity introduced by the summary writer. We find that most ambiguities in the summary claims are unintentionally introduced by the summarizer. 
Finally, we further motivate this problem by demonstrating that human-written summaries also exhibit ambiguities, indicating they are an inherent part of the task. 
Our contributions include:
\begin{enumerate}
\item An extended task definition of faithfulness in narrative summarization that allows for ambiguity in claim semantics.
\item Human-annotated subjectivity labels and human-written summaries for the StorySumm dataset to instantiate this task.
\item A new rewriting-based evaluation method, the Ambiguity Rewrite Metric (ARM), 
as an automatic evaluation for this task.
\end{enumerate}



\section{Background}
\label{sec:background}


Narrative summarization involves ambiguity and subjectivity \citep{subbiah2024reading}. \textit{Ambiguities} are phrases in the story or summary that have different interpretations. For example, in the introductory example in Figure \ref{fig:introexample}, the story implies an alien abduction without specifying it, leaving this interpretation ambiguous. Stories intentionally use ambiguity. Summaries can unintentionally introduce \textit{subjectivity} through resolving ambiguity with explicit interpretation or using paraphrase with slightly different semantics. When evaluating the faithfulness of summaries, using only a binary faithfulness label can lead to disagreement due to the subjectivity involved \citep{subbiah2024storysumm}. Importantly, this disagreement is beyond annotator error as the annotators have legitimate reasons for their choices of labels. 

 Instead of assigning one binary label to a summary or claim, we consider the following subtasks in evaluating narrative summaries, similar to \cite{wadhwa-etal-2024-learning-refine}:
\begin{enumerate}
\item \textbf{Detecting} ambiguities in a summary claim. In Figure \ref{fig:introexample}, the red circle indicates the ambiguous word ``aliens'' in the summary claim.
\item \textbf{Fixing} ambiguities in summary claims by producing rewrites of the claims. In Figure \ref{fig:introexample}, the rewrite changes the wording of the abduction to match the story.
\item \textbf{Explaining} what the ambiguity is with an explanation $E$. In Figure \ref{fig:introexample}, the explanation indicates the difference between the original summary claim and the story.
\end{enumerate}


We hypothesize that fixing ambiguities in summary claims 
can reduce their subjectivity and produce more objectively faithful claims. Approaching evaluation in this way therefore focuses on measuring how far claims are from objective faithfulness rather than trying to assign an objective label up front. We therefore propose using LLM rewrites as an automatic method of evaluation, the Ambiguity Rewrite Metric (ARM). 
This type of evaluation has been shown useful when executed by humans but we demonstrate that LLM-generated rewrites are an effective evaluation method \citep{nanba2004comparison,liu2022improving, yao2023improving}.



\section{Methods}



We consider the setting of a narrative $D$ which is summarized into a summary $S$. An LLM for this task places a distribution $p(S \mid D)$ and we consider sampling predicted summaries $\hat{S} \sim p(S \mid D)$. Each $S$ can be viewed as a collection of claims $(s_1,\ldots,s_n)$, which can be computed through a decomposition process. In our case, we consider each sentence in a summary as a claim. Instead of seeking a binary label, we consider each claim to have one of four faithfulness statuses: $f(s_i) \in \{\mathrm{supported}, \mathrm{unsupported}, \mathrm{ambiguous}, \mathrm{N/A}\}$. N/A claims are those which just provide commentary on the story and are meant to be subjective interpretation by nature (e.g., \textit{Overall, this is a story about love and how it overcomes obstacles}).
We primarily focus on evaluating supported, unsupported, and ambiguous claims.

Objective claims, those which are not ambiguous, are optimal from an evaluation perspective because annotators are able to agree on their faithfulness in relation to a story. Our method leverages this insight in evaluation.

\subsection{Claim rewriting}
\label{sec:rewriting}

We use a rewrite model $M$ to rewrite each claim $s$, producing a rewrite $r$ and optionally an explanation of the rewrite $E$ consisting of individual points, such that $M(s) = (r, E)$. In practice, we use an LLM as $M$ (with $\textit{temperature}=0$), prompting it to rewrite claims with ambiguities or unfaithful details 
or just repeat the original claim wording if there are no issues.


 Rewrites provide three quantitative feedback signals for automatic evaluation: 
\begin{enumerate}
    \item Whether $r = s$, which provides a binary label with 1 indicating $s$ does not contain ambiguities and 0 indicating it does.
    \item The edit distance between $r$ and $s$, which indicates how much the rewriting process changed in $s$ to address the ambiguities.
    \item The number of points discussed in the explanation, $|E|$, which indicates the number of ambiguities addressed.
\end{enumerate}
Rewrites additionally provide qualitative feedback: 
(1) the wording changes between $r$ and $s$ clearly indicate which phrasing is ambiguous in $s$ and how to correct this, and (2) the explanation $E$ explains in natural language the issues with $s$.

\subsection{Using rewrites for evaluation}
We can use the benefits of rewrites discussed in Section \ref{sec:rewriting} to address the evaluation subtasks introduced in Section \ref{sec:background}. For \textbf{detecting} ambiguities, we can use the binary label $r=s$ to detect if a claim $s$ contains ambiguities. The qualitative feedback provided by $r$ additionally identifies what these ambiguities are. The rewrite $r$ then \textbf{fixes} the ambiguities in $s$. In this way, rewrites elegantly accomplish multiple tasks in one step. Finally, the explanation $E$ \textbf{explains} what is ambiguous in $s$ that is addressed by $r$. 

The central question we ask is, \textit{does resolving ambiguities make claims more objectively faithful to humans?} If so, rewriting claims can serve as an interpretable evaluation of where a claim lies on the ambiguity spectrum in relation to objective faithfulness. 
When we use $r=s$ as a binary label for detecting ambiguities, we refer to this as the Ambiguity Rewrite Metric (\textbf{ARM}). When we use the rewrites as qualitative feedback, we refer to them as just the \textbf{rewrites}. 


\section{Experimental Setup}

\begin{figure}
\centering \small
\begin{tikzpicture}
    \begin{axis}[
        ybar,
        bar width=12pt,
        width=\columnwidth,
        height=4cm,
        symbolic x coords={Faithful, Unfaithful},
        xtick=data,
        ytick={0,100, 200, 300, 400, 500},
        ymin=0,
        ymax=500,
        xlabel={ },
        ylabel={\# of Claims},
        legend pos=north west,
        ymajorgrids=true,
        enlarge x limits=0.5,
        legend pos = north east,
        major tick length=0pt,
        legend style={nodes={scale=0.8, transform shape}}
    ]
        \addplot coordinates {(Faithful,444) (Unfaithful,20)};
        \addplot coordinates {(Faithful,37) (Unfaithful,66)};
        \legend{Objective, Subjective}
    \end{axis}
\end{tikzpicture}
\caption{A breakdown of our subjectivity labels by the original faithfulness labels in StorySumm, showing substantial overlap in subjective and unfaithful labels.} 
\label{fig:subjlabs}
\end{figure}

\subsection{Dataset}
We use the StorySumm dataset \citep{subbiah2024storysumm} to test our task definition and evaluation method. StorySumm consists of 32 English stories collected from amateur writing subreddits, and each summarized by three GPT 
or Claude 
series LLMs. In total, there are 96 summaries, consisting of 568 sentences. 
Each summary and sentence is labeled as faithful, unfaithful, or N/A, relative to the story. There are labels from multiple annotators along with written explanations of unfaithful labels.

\paragraph{Subjectivity annotation} 
Two of the authors of this paper assign a new label of objective or subjective for each claim $s$ in the dataset. They first read through annotator disagreements and explanations to identify a set of ambiguity types in the reasons why claims are ambiguous. They then code each claim with these types, and finally adjudicate any disagreements. Subjective claims are considered any claims assigned one of these types; the breakdown by types is discussed in Section \ref{sec:ambigtypes}. 
 \textit{Objective} claims can be faithful or unfaithful. 
In Figure \ref{fig:subjlabs}, we see the breakdown of subjectivity labels for faithful vs. unfaithful claims. Most of the unfaithful claims are also labeled as subjective. This overlap indicates that subjectivity is a challenging part of labeling faithfulness in claims.

\paragraph{Establishing subjectivity in this task} 
To demonstrate whether subjectivity is an inherent part of this task or is introduced by LLMs in their summaries, we compare the LLM summaries against \textbf{human-written summaries}.\footnote{We release the human summaries and subjectivity annotations at: \url{https://github.com/melaniesubbiah/storysumm}} 
We recruit graduate students in a computer science department 
to write these summaries who we trust to complete the task without LLM assistance. We collect five summaries from three students, resulting in 15 summaries, each for a different story in StorySumm, and 108 claims to evaluate. 

\label{sec:syntheticdata}
To validate our definition of subjectivity, we also test whether inserting or removing the types of ambiguities we look for affects annotator agreement on faithfulness labels in the way we expect. As part of our subjectivity annotation, we identify four types of ambiguities (discussed in detail in Section \ref{sec:ambigtypes}). We generate \textbf{synthetic claims} using Claude-3.5 with prompts designed to produce subjective variants of objective claims in StorySumm and objective variants of subjective claims (see prompts in Appendix \ref{sec:appsynthprompts}). Each prompt introduces or corrects one of the ambiguity types we identify in our annotation.  

\subsection{Human studies with Upwork annotators}

We perform several human studies, each using three Upwork annotators who pass a pilot screening. 
In each case, we show annotators a story and summary from StorySumm and ask them questions about a specific claim in the summary (see Appendix Figures \ref{fig:appinstructions} and \ref{fig:apptasks} for the study interfaces). We 
use annotator disagreement on a faithfulness label for a claim as a proxy for subjectivity. While some disagreement will always arise from annotator error, significantly greater disagreement in comparing claim settings indicates more subjectivity. 

We first study whether ambiguities are essential to address in evaluation:
\begin{itemize}
\item\textbf{Are ambiguities inherent to this task?} For the human-written summaries, we ask whether each claim in each summary is faithful to the story. We compare the level of disagreement between the three annotators on these human-written summaries vs. on the original LLM-generated summaries for these stories. This study indicates whether humans also introduce subjective claims in narrative summarization.  
\item\textbf{Do ambiguities impact claim subjectivity?} Using the synthetic 
claims, we create spliced 
summaries that are assembled by randomly selecting an objective or subjective variant for each claim in a summary. We ask whether each claim in this synthetic summary is faithful to the story and compare annotator disagreement on claims we expect to be objective vs. subjective. This study indicates whether the ambiguities we identify in claims have a measurable impact on claim subjectivity.
\end{itemize}
In early experimentation, we find that Claude-3.5-Sonnet 
is a strong rewrite model $M$ for this task, so we use it with the subjectivity-targeted prompt shown in Appendix \ref{sec:appbaselineprompts} for a detailed comparison to human judgments. To compare rewrites with human judgments, we use a three stage evaluation for each rewritten claim:
\begin{itemize}
\item \textbf{Does claim rewriting reduce subjectivity?} We randomly show either the original or rewrite in the summary and ask whether that claim is faithful to the story. If we see greater average agreement between annotators on rewritten claims than original claims, we know rewrites make the claims more clear and objective to evaluate.
\item \textbf{Do rewrites improve claims?} We then show whichever version of the claim was not presented in the summary as an alternate and ask which version is better. We can observe whether rewrites are significantly preferred over original claims.
\item \textbf{Are explanations of rewrites meaningful?} For rewrites which annotators prefer, we parse the LLM-generated explanation for the rewrite into individual points (see prompt in Appendix \ref{sec:appexpprompt}). We ask annotators to judge whether each point is important to their choice for why the rewrite is better. If the explanation is accurate in relation to the story and claim, the annotator can label it as IMPORTANT to their preference or NEUTRAL to their preference, and if the explanation is inaccurate, they can label it as WRONG. These annotations indicate whether explanations discuss meaningful changes in the claims. 
\end{itemize}

 We compare other rewrite models quantitatively but cannot perform a human study for all models. See Appendix \ref{sec:appannovalid} for details on how we validate this annotation format.

\subsection{Metrics}


For annotator agreement, we compute the percent of claims for which all three annotators assign the same faithfulness label. For word-level edit distance between claims, we tokenize each sentence, lowercase and stem words, remove whitespace and then compute the Levenshtein distance \citep{navarro2001guided} using these individual words as the atomic units. We use balanced accuracy and F1-macro scores as measures of classification accuracy on imbalanced datasests for evaluating detection accuracy for subjectivity.

\begin{figure*}[t]
\centering
    \includegraphics[width=\linewidth]{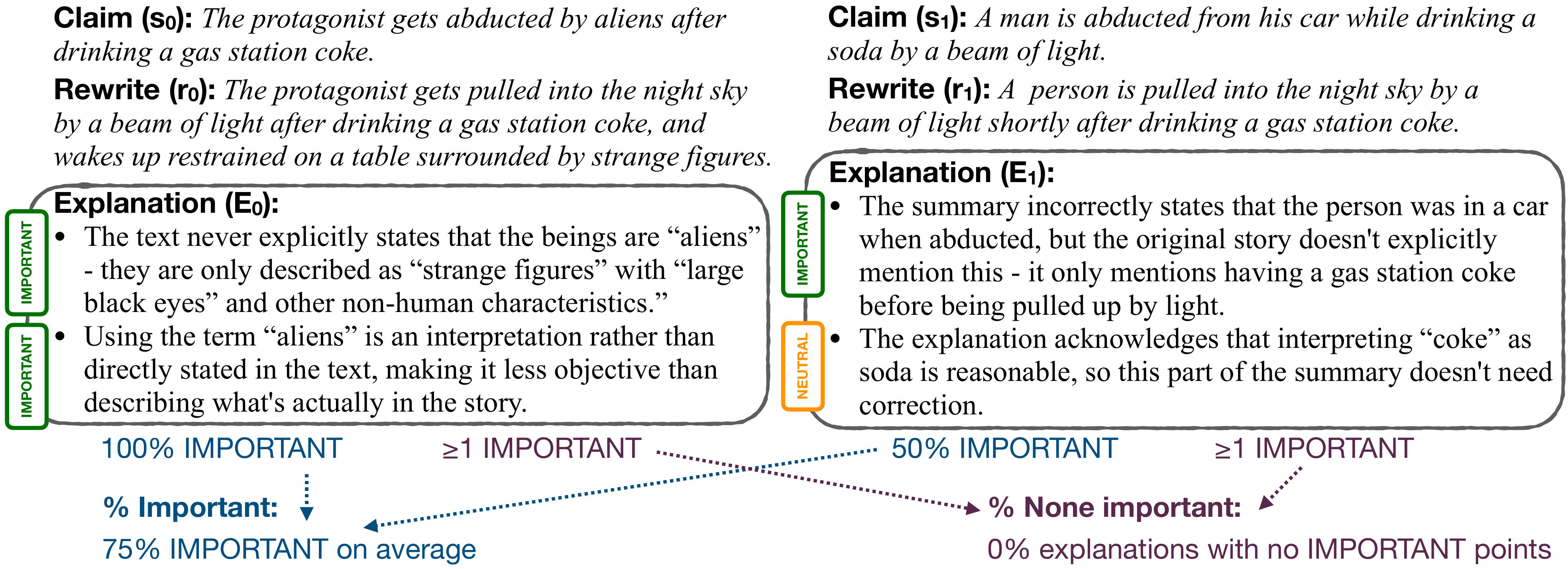}
\caption{Examples of computing the metrics used for evaluating explanations, computing 75\% for \textit{\% important} and 0\% for \textit{\% none important}.
}
    \label{fig:metrics}
\end{figure*}

For evaluating explanations, we use metrics defined as follows. Let $l$ be an individual label for an explanation point, which could take the value IMPORTANT, NEUTRAL, or WRONG. Then $E$ is the set of labels $l$ corresponding to one full explanation, and $R$ is the set of $E$ corresponding to all the rewrites in the dataset. We then define \textit{\% important} as the macro average of the fraction of points labeled important:
\begin{equation*}
\frac{1}{|R|}\sum_{E \in R}{\frac{1}{|E|}\sum_{l \in E}{\mathbf{1}[l = \text{IMPORTANT}]}}
\end{equation*}
and \textit{\% none important} 
as the fraction of totally not important explanations:
\begin{equation*}
1- \frac{1}{|R|}\sum_{E \in R}{\mathbf{1}[\text{IMPORTANT} \in E]}
\end{equation*}
We can similarly calculate \textit{\% wrong} and \textit{\% none wrong} by swapping WRONG for IMPORTANT in these equations. Intuitively, \textit{\% important} averages across the dataset the fraction of explanation points that are labeled IMPORTANT for a rewrite. \textit{\% none important} is the fraction of rewrites with no explanation points labeled IMPORTANT. This difference is shown visually in Figure \ref{fig:metrics}. 

For statistical significance, we report a bootstrap significance test with 10,000 trials. 

\subsection{Baselines and rewrite models}
\label{sec:baselines}
To compare against rewrites, we use standard methods of prompting and finetuning LLMs as baselines for detecting subjective claims. We use GPT-4 \citep{openai2024gpt4technicalreport} and Claude-3.5-Sonnet. We try \textbf{zero-shot} and \textbf{few-shot prompting}, asking whether a claim is objective. We try a \textbf{self-consistency} method \citep{wang2022self} of sampling three different zero-shot CoT answers \citep{wei2022chain} for whether or not a claim is faithful with temperature 0.7. Full prompts are in Appendix \ref{sec:appbaselineprompts}.

We also compare against a \textbf{fine-tuned} model. We use a Llama-3.1-8B-Instruct model \citep{grattafiori2024llama} and the synthetically generated data of objective and subjective claims discussed in Section \ref{sec:syntheticdata}. 
This method results in a dataset of about 2k claims that we can finetune on. 
We apply LoRA \citep{hu2022lora} with a rank of 64, alpha set to 64, and a dropout rate of 0.05. For hyperparameters, we use a learning rate of 5e-5, a batch size of 8, gradient accumulation over 2 steps, and train for 1 epoch using two A100 GPUs.

To compare different rewrite models, we use GPT-4o and Claude-3.5-Sonnet with a standard rewriting prompt targeting subjectivity and unfaithfulness. We additionally compare this prompt against other variants targeting only subjectivity or only unfaithfulness (see Appendix \ref{sec:appbaselineprompts} for all prompts). 


\section{Results}

Using the models and methods discussed, we establish subjectivity as part of narrative summarization and identify why claims are subjective. We demonstrate rewriting effectively detects subjectivity and corrects it in alignment with human judgments.   

\subsection{Does subjectivity play a role in evaluation of this task?}

We compare whether human-written summaries exhibit different levels of subjectivity than LLM-written summaries. 
As shown in Table \ref{tab:agreecompare}, we find that all three annotators agree on 76\% of the human-written claims compared to 73\% of the LLM-generated claims in StorySumm. This is not a statistically significant difference ($p$: 0.23), indicating that \textbf{ambiguous claims are an inherent challenge in evaluating narrative summaries}, and are not specific to LLMs.

\begin{table}
\centering \small
\begin{tabular}{ll}
\toprule
& \textbf{\% Agree}\\
\textbf{Summary writer}:&\\\cmidrule{1-1}
Human & 75.93\\
LLM & 72.74\\\midrule
\textbf{Synthetic claim type}: & \\\cmidrule{1-1}
Objective & $\textbf{55.56}^{**}$\\
Subjective & 26.56\\\bottomrule
\end{tabular}
\caption{For each type of claim, we report the percent of those claims for which all three annotators agree on the faithfulness label.} 
\label{tab:agreecompare}
\end{table}

We compare whether claims generated to introduce or address 
the ambiguities we identify exhibit greater subjectivity or objectivity respectively. As shown in Table \ref{tab:agreecompare}, 
27\% of synthetically generated subjective claims produce annotator agreement compared to 56\% of synthetically generated objective claims. This difference is statistically significant 
($p: 2e$-4), demonstrating \textbf{the ambiguities we identify lead to annotator disagreement}. Given that these claims are synthetically generated, we do not see perfect agreement or disagreement for either category. Many of the objective claims which still produce disagreement are due to annotator error, and the subjective claims with agreement are generally due to synthetic data generation introducing an easily identifiable inconsistency.

\subsection{Can rewrites detect subjectivity?}

\begin{table}[t]
\centering \small
\begin{tabular}{lcc}
\toprule
\textbf{Method} & \textbf{Bal. acc.} & \textbf{F1-macro}\\\midrule\midrule
Llama-3.1-8B finetuned & 49.62 & 0.50 \\\midrule
Claude-3.5 zero-shot & 63.85 &  0.61\\
Claude-3.5 few-shot & 66.21 & \textbf{0.62}\\
Claude-3.5 self-consistency & 50.42 & 0.50\\
Claude-3.5 ARM (ours) & \textbf{69.15} & 0.53\\\midrule
GPT-4 zero-shot & 58.04 & 0.57  \\
GPT-4 few-shot & 58.94 & 0.56\\
GPT-4 self-consistency & 57.73 & \textbf{0.59}\\
GPT-4 ARM (ours) & \textbf{63.55} & 0.58\\\bottomrule
\end{tabular}
\caption{Performance of rewrites with different models relative to baseline methods for subjectivity detection against our subjectivity labels on StorySumm. We report balanced accuracy and F1-macro scores.}
\label{tab:detectionresults}
\end{table}

The ARM metric labels claims which get rewritten as subjective 
and labels them objective otherwise. We compare how well ARM detects subjectivity relative to the baselines detailed in Section \ref{sec:baselines} in Table \ref{tab:detectionresults}. 
We 
find that finetuning a detection model on the synthetic dataset does not transfer effectively to the original claims. In zero-shot and few-shot settings, we see that Claude is overall a stronger model for detecting subjectivity than GPT-4. Self-consistency does not work well for Claude but has comparable performance to other methods for GPT-4. ARM is a strong method for detecting subjectivity across both models, and the strongest when looking at balanced accuracy. Overall, 
\textbf{ARM is a stronger or comparable method to other baselines for binary evaluation while providing the added benefits of the rewrite itself and explanation}. We report additional results with alternate prompts and against faithfulness labels next and observe that ARM is fairly consistent across different prompts.


\subsection{Does the rewrite model and prompt affect results?}
\begin{table*}[ht!]
\centering \small
\begin{tabular}{lccc|cc}\toprule
\bf Method & \bf Subj. & \bf Unfaith. & \bf Subj. $\lor$ unfaith. & \bf \# Rewrites & \bf Avg. edit dist.\\\midrule
Claude subj. & 67.89 & 67.42 & 69.00 & 246 & 0.40\\
Claude both & \textbf{69.15} & 68.05 & \textbf{69.96} & 341 & 0.61 \\
Claude unfaith. & 66.7 & \textbf{68.21} & 63.56 & 218 & 0.46 \\
GPT-4 subj. & 53.95 & 55.42 & 54.61 & 486 & 0.17\\
GPT-4 both & 63.55 & 64.95 & 63.39 & 203 & 0.22 \\
GPT-4 unfaith. & 64.01 & 63.09 & 63.56 & 390 & 0.21 \\\bottomrule
\end{tabular}
\caption{We report the balanced accuracy scores for different prompts and models against the subjectivity labels (Subj.), faithfulness labels (Unfaith.) and a combination of both label sets that looks for claims which are subjective or unfaithful (Subj. $\lor$ Unfaith.). Subj. and unfaith. indicate the prompts targeting subjectivity and unfaithfulness respectively, and the ``both" method indicates the prompt shown in Section \ref{sec:rewriting} which targets both subjectivity and unfaithfulness. We also report the number of claims which are rewritten and the average edit distance of rewrites relative to the originals.} 
\label{tab:rewriteresults}
\end{table*}

We 
assess how sensitive rewrites are to prompting method and model choice in Table \ref{tab:rewriteresults}. We evaluate detection against subjectivity labels, faithfulness labels, and claims that are subjective or unfaithful. We use the rewriting prompts shown in Appendix \ref{sec:appbaselineprompts}. 
We see that Claude is a stronger model for rewriting in terms of binary classification of subjectivity or faithfulness. GPT-4 provides the benefit of making more minimal changes to claims
as its rewrites have less than half the edit distance relative to Claude. Overall, \textbf{the LLM used for rewriting matters more than the prompt} with the exception of the subjectivity-targeted prompt for GPT-4. This setting produces the most rewrites by far which leads to its poor performance. 

\subsection{Do rewrites align with human judgments of faithfulness?}

\begin{table}
\centering \small
\begin{tabular}{lcc}\toprule
\bf Metric & \bf Original claims & \bf Rewrites\\\midrule
Agreement & 36.36 & 57.45*\phantom{0}\\
Faithful & 20.45 & 89.36** \\
Preferred & 23.08 & 76.92**\\\bottomrule
\end{tabular}
\caption{The percent of each claim type that meets each condition. Agreement requires all three annotators agree on the label. Faithful requires at least two of the annotators label the claim faithful. Preferred requires that claim variant was preferred over the other. ($p\leq.05: *$, $p\leq.001: **$)}
\label{tab:annoresults}
\end{table}

We take the subjectivity-focused rewrites using Claude (prompt in Appendix \ref{sec:appbaselineprompts}) and test whether its edits align with human judgments and preferences. In Table \ref{tab:annoresults}, we see that rewrites exhibit statistically significant gains over original claims in annotator agreement and faithfulness. When human judges are asked whether they prefer the original claim versus the rewritten claim, they prefer the rewrites 77\% of the time. 

In Table \ref{tab:expresults}, we report the quality of explanations. 
We find that on average 69\% of explanation points are labeled as IMPORTANT by majority vote. 99\% of explanations have no points labeled WRONG by majority vote. Finally, only 5\% of explanations have no points considered important for an individual annotator. These percentages show that the vast majority of explanations and changes made in the rewrites are either important to fix or neutral. We note that the explanation parse often repeats the same point multiple times (see example explanations in Figure \ref{fig:metrics}), which is why we consider both \textit{\% important} and \textit{\% wrong}. 

These numbers indicate that \textbf{rewrites significantly improve claims in objectivity and faithfulness and using them in conjunction with their explanations is a meaningful evaluation signal}.

\begin{table*}[h!]
\centering \small
\begin{tabular}{lcccc}
\toprule
           & \multicolumn{2}{c}{\bf Important} & \multicolumn{2}{c}{\bf Wrong} \\\cmidrule(lr){2-3}\cmidrule(lr){4-5}
\textbf{Method}     & \% Important       & \% None important      & \% Wrong    & \% None wrong    \\\midrule
Individual & 84.11 & \phantom{0}5.29 & 7.52 & 87.83\\
Maj. vote & 68.72 & 24.39 & 0.56 & 98.78\\\bottomrule
\end{tabular}
\caption{Percentages for explanation annotations (metrics described in Figure \ref{fig:metrics}).}
\label{tab:expresults}
\end{table*}


\subsection{Qualitative analysis of ambiguities}
\label{sec:ambigtypes}


We uncover specific reasons why claims are ambiguous and analyze whether there are specific types of ambiguities that are harder for rewrites to detect. 
Two of the authors of this paper perform inductive thematic analysis \citep{bowman2023using} with adjudication to identify a taxonomy for why claims are subjective (see Appendix \ref{sec:appambiganalysis} for more details). 
Using this taxonomy, they revisit each subjective claim and label it with an ambiguity code. The taxonomy identifies four ambiguity types:

\begin{enumerate}
\item \textbf{Wording}: A word or phrase is used in the summary which has overlapping meaning with the wording used in the story but also lends itself to other interpretations. Depending on someone's interpretation, the two meanings may not fully overlap. 
\item \textbf{Detail}: A very minor detail is assumed in the summary which is not explicitly stated in the story. Many people find this to be a reasonable assumption and therefore faithful, while others view it as unfaithful.
\item \textbf{Causation}: The summary skips important causal details for an event. Some people find this to be a reasonable abbreviation while others feel it fundamentally changes one's understanding of what happened.
\item \textbf{Explicit}: The summary makes explicit details that are intentionally left ambiguous or only implied in the story. Some people like this interpretive jump while others feel it misrepresents the nature of the story.
\end{enumerate}



In Figure \ref{fig:barcharts}, we see the most common ambiguity type in the LLM-generated subjective claims in StorySumm is \textbf{Type 1}
, meaning \textbf{many claims use vague or misleading wording}. This result is consistent with prior work \citep{subbiah2024reading, kim2024fables}. 
\textbf{Type 4} is the smallest category, indicating that most ambiguities are unintentionally introduced by the summary writer (\textbf{Types 1-3}). 
We also see that on average, there does not seem to be one type of ambiguity that is detected substantially more or less by the LLM rewrite metrics. We compare ARM's recall across the different ambiguity types but do not find that one type of claim is missed more often than others (see Appendix Figure \ref{fig:appbarcharts}).

\section{Related Work}

\begin{figure}
\centering \small
\begin{tikzpicture}
    \begin{axis}[
        ybar,
        bar width=10pt,
        width=\columnwidth,
        height=4cm,
        symbolic x coords={1, 2, 3, 4},
        xtick=data,
        ytick={0, 10, 20, 30, 40},
        ymin=0,
        ymax=40,
        xlabel={Ambiguity type},
        ylabel={\# of Subjective claims},
        ymajorgrids=true,
        enlarge x limits=0.3,
        legend pos = north east,
        major tick length=0pt
    ]
        \addplot coordinates {(1, 37) (2, 22) (3, 17) (4, 10)};
    \end{axis}
\end{tikzpicture}
\caption{A breakdown of how many times each ambiguity type is labeled in the StorySumm LLM-generated summaries. }
\label{fig:barcharts}
\end{figure}

\textbf{Ambiguity in evaluation} \quad Ambiguity has been studied in natural language entailment \citep{pavlick2019inherent, nie2020can, jiang-marneffe-2022-investigating}, a related task to summary faithfulness evaluation. Other work has studied how to improve or manage annotator disagreement \citep{uma2021learning, krishna2023longeval, min2023factscore}. Most similar to ours are \citet{koupaee2025faithful, mishra2024finegrained, ramprasad-etal-2024-analyzing}, which include various types of ambiguities in evaluating meeting summaries, real-world QA, and dialogue summaries respectively. Their taxonomies of ambiguities support our findings and by working with narratives, we incorporate the added challenge of intentional ambiguity in the source text.

\textbf{Edits as feedback} \quad Recent work has explored the use of natural language feedback to improve model outputs in different settings, such as math reasoning \citep{madaan2023selfrefine, xu-etal-2024-chatglm}  and summarization \citep{liu2022improving, zhang-etal-2023-summit}. This feedback can be human feedback \citep{madaan2023selfrefine} or automatic feedback from models \citep{gao-etal-2023-rarr, selfee2023, wadhwa-etal-2024-learning-refine, xu-etal-2024-llmrefine}. While most of these methods follow a critique-and-refine framework to improve models' outputs, we propose a rewriting-based evaluation method that leverages edits themselves as the feedback signal.

\textbf{Evaluating narrative summaries} \quad Prior work has introduced datasets and methods for evaluating narrative summaries. Early work provided reference summaries from online study guide and TV episode synopsis websites to enable reference-based evaluation metrics \citep{ladhak2020exploring, kryscinski2021booksum, chen2021summscreen}. As LLMs have trained on more and more online data, more recent work has evaluated LLMs as evaluators on recently published books or unpublished work \citep{chang2024booookscore, kim2024fables, subbiah2024reading, karpinska2024one}.

\section{Conclusion}
In this work, we expand the considerations for evaluating faithfulness in narrative summarization to include ambiguity and subjectivity. We propose using rewrites as automatic evaluation for summary claims and demonstrate their efectiveness on the StorySumm dataset. We release additional labels for StorySumm for claim subjectivity and ambiguity types. In the future, we hope this evaluation methodology can be tested on other challenging evaluation tasks in the humanities. We believe rewriting-based evaluation could also be used in RL training for improving reasoning  about implicit meaning in the humanities.



\section*{Ethics Statement}
There are no risks involved with this work as it explores summarization evaluation with publicly available data. We follow protocol approved by Columbia University IRB protocol AAAS4051 for the human annotation work in this study. We use AI assistants to answer questions about coding for this work, but do not use them for any part of research design or paper writing. The StorySumm dataset is available for research use and our use is in line with this purpose. One of the authors, Melanie Subbiah, holds an equity interest in OpenAI.

\section*{Limitations}
The StorySumm dataset is relatively small which limits the generalization of our conclusions. However, we are still able to see statistically significant results with this dataset. Using human evaluation is a thorough and rich source of feedback but can limit reproducibility of results since different annotators will perform the task slightly differently. Finally, given that we study subjectivity in this work, there is inherent subjectivity involved in the task which can further limit the reproducibility of results. Despite this, we feel it is important to try to formalize and make progress on areas of evaluation that contain grey areas.

\section*{Acknowledgments}
We would like to express our gratitude to the Upwork workers who contributed annotations for this work. Additionally, we would like to thank our reviewers for their thoughtful feedback. This work is supported by the funds provided by several organizations: the Columbia Amazon CAIT PhD Fellowship, Northrup Grumman, the National Science Foundation, DoD OUSD (R\&E) under Cooperative Agreement PHY-2229929 (The NSF AI Institute for Artificial and Natural Intelligence), and Good Systems,\footnote{https://goodsystems.utexas.edu/} a UT Austin Grand Challenge to develop responsible AI technologies.

\bibliography{custom}

\begin{thebibliography}{41}
\providecommand{\natexlab}[1]{#1}

\bibitem[{Bowman et~al.(2023)Bowman, Nadal, Morrissey, Thieme, and Doherty}]{bowman2023using}
Robert Bowman, Camille Nadal, Kellie Morrissey, Anja Thieme, and Gavin Doherty. 2023.
\newblock \href {https://doi.org/10.1145/3544548.3581203} {{Using Thematic Analysis in Healthcare HCI at CHI: A Scoping Review}}.
\newblock In \emph{Proceedings of the 2023 CHI Conference on Human Factors in Computing Systems}, CHI '23, New York, NY, USA. Association for Computing Machinery.

\bibitem[{Chang et~al.(2024)Chang, Lo, Goyal, and Iyyer}]{chang2024booookscore}
Yapei Chang, Kyle Lo, Tanya Goyal, and Mohit Iyyer. 2024.
\newblock {BooookScore: A systematic exploration of book-length summarization in the era of {LLM}s}.
\newblock In \emph{The Twelfth International Conference on Learning Representations}.

\bibitem[{Chen et~al.(2022)Chen, Chu, Wiseman, and Gimpel}]{chen2021summscreen}
Mingda Chen, Zewei Chu, Sam Wiseman, and Kevin Gimpel. 2022.
\newblock \href {https://doi.org/10.18653/v1/2022.acl-long.589} {{{S}umm{S}creen: A Dataset for Abstractive Screenplay Summarization}}.
\newblock In \emph{Proceedings of the 60th Annual Meeting of the Association for Computational Linguistics (Volume 1: Long Papers)}, pages 8602--8615, Dublin, Ireland. Association for Computational Linguistics.

\bibitem[{Chen et~al.(2023)Chen, Zhao, Zhang, Chern, Gao, Liu, and He}]{chen2023felm}
Shiqi Chen, Yiran Zhao, Jinghan Zhang, I-Chun Chern, Siyang Gao, Pengfei Liu, and Junxian He. 2023.
\newblock {FELM: Benchmarking Factuality Evaluation of Large Language Models}.
\newblock In \emph{Thirty-seventh Conference on Neural Information Processing Systems Datasets and Benchmarks Track}.

\bibitem[{Durmus et~al.(2020)Durmus, He, and Diab}]{durmus2020feqa}
Esin Durmus, He~He, and Mona Diab. 2020.
\newblock \href {https://doi.org/10.18653/v1/2020.acl-main.454} {{{FEQA}: A Question Answering Evaluation Framework for Faithfulness Assessment in Abstractive Summarization}}.
\newblock In \emph{Proceedings of the 58th Annual Meeting of the Association for Computational Linguistics}, pages 5055--5070, Online. Association for Computational Linguistics.

\bibitem[{Fabbri et~al.(2021)Fabbri, Kry{\'s}ci{\'n}ski, McCann, Xiong, Socher, and Radev}]{fabbri-etal-2021-summeval}
Alexander~R. Fabbri, Wojciech Kry{\'s}ci{\'n}ski, Bryan McCann, Caiming Xiong, Richard Socher, and Dragomir Radev. 2021.
\newblock \href {https://doi.org/10.1162/tacl_a_00373} {{{S}umm{E}val: Re-evaluating Summarization Evaluation}}.
\newblock \emph{Transactions of the Association for Computational Linguistics}, 9:391--409.

\bibitem[{Gao et~al.(2023)Gao, Dai, Pasupat, Chen, Chaganty, Fan, Zhao, Lao, Lee, Juan, and Guu}]{gao-etal-2023-rarr}
Luyu Gao, Zhuyun Dai, Panupong Pasupat, Anthony Chen, Arun~Tejasvi Chaganty, Yicheng Fan, Vincent Zhao, Ni~Lao, Hongrae Lee, Da-Cheng Juan, and Kelvin Guu. 2023.
\newblock \href {https://doi.org/10.18653/v1/2023.acl-long.910} {{{RARR}: Researching and Revising What Language Models Say, Using Language Models}}.
\newblock In \emph{Proceedings of the 61st Annual Meeting of the Association for Computational Linguistics (Volume 1: Long Papers)}, pages 16477--16508, Toronto, Canada. Association for Computational Linguistics.

\bibitem[{Grattafiori et~al.(2024)Grattafiori, Dubey, Jauhri, Pandey, Kadian, Al-Dahle, Letman, Mathur, Schelten, Vaughan et~al.}]{grattafiori2024llama}
Aaron Grattafiori, Abhimanyu Dubey, Abhinav Jauhri, Abhinav Pandey, Abhishek Kadian, Ahmad Al-Dahle, Aiesha Letman, Akhil Mathur, Alan Schelten, Alex Vaughan, and 1 others. 2024.
\newblock {The Llama 3 Herd of Models}.
\newblock \emph{arXiv}.

\bibitem[{Hu et~al.(2022)Hu, Shen, Wallis, Allen-Zhu, Li, Wang, Wang, Chen et~al.}]{hu2022lora}
Edward~J Hu, Yelong Shen, Phillip Wallis, Zeyuan Allen-Zhu, Yuanzhi Li, Shean Wang, Lu~Wang, Weizhu Chen, and 1 others. 2022.
\newblock {Lora: Low-rank adaptation of large language models}.
\newblock \emph{International Conference on Learning Representations (ICLR)}.

\bibitem[{Jiang and de~Marneffe(2022)}]{jiang-marneffe-2022-investigating}
Nan-Jiang Jiang and Marie-Catherine de~Marneffe. 2022.
\newblock \href {https://doi.org/10.1162/tacl_a_00523} {{Investigating Reasons for Disagreement in Natural Language Inference}}.
\newblock \emph{Transactions of the Association for Computational Linguistics}, 10:1357--1374.

\bibitem[{Karpinska et~al.(2024)Karpinska, Thai, Lo, Goyal, and Iyyer}]{karpinska2024one}
Marzena Karpinska, Katherine Thai, Kyle Lo, Tanya Goyal, and Mohit Iyyer. 2024.
\newblock \href {https://doi.org/10.18653/v1/2024.emnlp-main.948} {{One Thousand and One Pairs: A {\textquotedblleft}novel{\textquotedblright} challenge for long-context language models}}.
\newblock In \emph{Proceedings of the 2024 Conference on Empirical Methods in Natural Language Processing}, pages 17048--17085, Miami, Florida, USA. Association for Computational Linguistics.

\bibitem[{Kim et~al.(2024)Kim, Chang, Karpinska, Garimella, Manjunatha, Lo, Goyal, and Iyyer}]{kim2024fables}
Yekyung Kim, Yapei Chang, Marzena Karpinska, Aparna Garimella, Varun Manjunatha, Kyle Lo, Tanya Goyal, and Mohit Iyyer. 2024.
\newblock {FABLES: Evaluating faithfulness and content selection in book-length summarization}.
\newblock In \emph{Conference on Language Modeling (COLM)}.

\bibitem[{Koupaee et~al.(2025)Koupaee, Vincent, Mansour, Shalyminov, He, Song, Shu, He, Nian, Wong et~al.}]{koupaee2025faithful}
Mahnaz Koupaee, Jake~W Vincent, Saab Mansour, Igor Shalyminov, Han He, Hwanjun Song, Raphael Shu, Jianfeng He, Yi~Nian, Amy Wing-mei Wong, and 1 others. 2025.
\newblock {Faithful, Unfaithful or Ambiguous? Multi-Agent Debate with Initial Stance for Summary Evaluation}.
\newblock \emph{arXiv}.

\bibitem[{Krishna et~al.(2023)Krishna, Bransom, Kuehl, Iyyer, Dasigi, Cohan, and Lo}]{krishna2023longeval}
Kalpesh Krishna, Erin Bransom, Bailey Kuehl, Mohit Iyyer, Pradeep Dasigi, Arman Cohan, and Kyle Lo. 2023.
\newblock \href {https://doi.org/10.18653/v1/2023.eacl-main.121} {{{L}ong{E}val: Guidelines for Human Evaluation of Faithfulness in Long-form Summarization}}.
\newblock In \emph{Proceedings of the 17th Conference of the European Chapter of the Association for Computational Linguistics}, pages 1650--1669, Dubrovnik, Croatia. Association for Computational Linguistics.

\bibitem[{Kryscinski et~al.(2022)Kryscinski, Rajani, Agarwal, Xiong, and Radev}]{kryscinski2021booksum}
Wojciech Kryscinski, Nazneen Rajani, Divyansh Agarwal, Caiming Xiong, and Dragomir Radev. 2022.
\newblock \href {https://doi.org/10.18653/v1/2022.findings-emnlp.488} {{{BOOKSUM}: A Collection of Datasets for Long-form Narrative Summarization}}.
\newblock In \emph{Findings of the Association for Computational Linguistics: EMNLP 2022}, pages 6536--6558, Abu Dhabi, United Arab Emirates. Association for Computational Linguistics.

\bibitem[{Laban et~al.(2022)Laban, Schnabel, Bennett, and Hearst}]{laban2022summac}
Philippe Laban, Tobias Schnabel, Paul~N. Bennett, and Marti~A. Hearst. 2022.
\newblock \href {https://doi.org/10.1162/tacl_a_00453} {{{S}umma{C}: Re-Visiting {NLI}-based Models for Inconsistency Detection in Summarization}}.
\newblock \emph{Transactions of the Association for Computational Linguistics}, 10:163--177.

\bibitem[{Ladhak et~al.(2020)Ladhak, Li, Al-Onaizan, and McKeown}]{ladhak2020exploring}
Faisal Ladhak, Bryan Li, Yaser Al-Onaizan, and Kathleen McKeown. 2020.
\newblock \href {https://doi.org/10.18653/v1/2020.acl-main.453} {{Exploring Content Selection in Summarization of Novel Chapters}}.
\newblock In \emph{Proceedings of the 58th Annual Meeting of the Association for Computational Linguistics}, pages 5043--5054, Online. Association for Computational Linguistics.

\bibitem[{Liu et~al.(2023)Liu, Deb, Teruel, Halfaker, Radev, and Awadallah}]{liu2022improving}
Yixin Liu, Budhaditya Deb, Milagro Teruel, Aaron Halfaker, Dragomir Radev, and Ahmed~Hassan Awadallah. 2023.
\newblock \href {https://doi.org/10.18653/v1/2023.acl-long.844} {{On Improving Summarization Factual Consistency from Natural Language Feedback}}.
\newblock In \emph{Proceedings of the 61st Annual Meeting of the Association for Computational Linguistics (Volume 1: Long Papers)}, pages 15144--15161, Toronto, Canada. Association for Computational Linguistics.

\bibitem[{Madaan et~al.(2023)Madaan, Tandon, Gupta, Hallinan, Gao, Wiegreffe, Alon, Dziri, Prabhumoye, Yang, Gupta, Majumder, Hermann, Welleck, Yazdanbakhsh, and Clark}]{madaan2023selfrefine}
Aman Madaan, Niket Tandon, Prakhar Gupta, Skyler Hallinan, Luyu Gao, Sarah Wiegreffe, Uri Alon, Nouha Dziri, Shrimai Prabhumoye, Yiming Yang, Shashank Gupta, Bodhisattwa~Prasad Majumder, Katherine Hermann, Sean Welleck, Amir Yazdanbakhsh, and Peter Clark. 2023.
\newblock {Self-Refine: Iterative Refinement with Self-Feedback}.
\newblock In \emph{Thirty-seventh Conference on Neural Information Processing Systems}.

\bibitem[{Min et~al.(2023)Min, Krishna, Lyu, Lewis, Yih, Koh, Iyyer, Zettlemoyer, and Hajishirzi}]{min2023factscore}
Sewon Min, Kalpesh Krishna, Xinxi Lyu, Mike Lewis, Wen-tau Yih, Pang Koh, Mohit Iyyer, Luke Zettlemoyer, and Hannaneh Hajishirzi. 2023.
\newblock \href {https://doi.org/10.18653/v1/2023.emnlp-main.741} {{{FA}ct{S}core: Fine-grained Atomic Evaluation of Factual Precision in Long Form Text Generation}}.
\newblock In \emph{Proceedings of the 2023 Conference on Empirical Methods in Natural Language Processing}, pages 12076--12100, Singapore. Association for Computational Linguistics.

\bibitem[{Mishra et~al.(2024)Mishra, Asai, Balachandran, Wang, Neubig, Tsvetkov, and Hajishirzi}]{mishra2024finegrained}
Abhika Mishra, Akari Asai, Vidhisha Balachandran, Yizhong Wang, Graham Neubig, Yulia Tsvetkov, and Hannaneh Hajishirzi. 2024.
\newblock Fine-grained hallucination detection and editing for language models.
\newblock In \emph{First Conference on Language Modeling}.

\bibitem[{Nanba and Okumura(2004)}]{nanba2004comparison}
Hidetsugu Nanba and Manabu Okumura. 2004.
\newblock \href {https://aclanthology.org/L04-1116/} {{Comparison of Some Automatic and Manual Methods for Summary Evaluation Based on the Text Summarization Challenge 2}}.
\newblock In \emph{Proceedings of the Fourth International Conference on Language Resources and Evaluation ({LREC}`04)}, Lisbon, Portugal. European Language Resources Association (ELRA).

\bibitem[{Navarro(2001)}]{navarro2001guided}
Gonzalo Navarro. 2001.
\newblock \href {https://doi.org/10.1145/375360.375365} {{A guided tour to approximate string matching}}.
\newblock \emph{ACM Comput. Surv.}, 33(1):31–88.

\bibitem[{Nie et~al.(2020)Nie, Zhou, and Bansal}]{nie2020can}
Yixin Nie, Xiang Zhou, and Mohit Bansal. 2020.
\newblock \href {https://doi.org/10.18653/v1/2020.emnlp-main.734} {{What Can We Learn from Collective Human Opinions on Natural Language Inference Data?}}
\newblock In \emph{Proceedings of the 2020 Conference on Empirical Methods in Natural Language Processing (EMNLP)}, pages 9131--9143, Online. Association for Computational Linguistics.

\bibitem[{OpenAI(2024)}]{openai2024gpt4technicalreport}
OpenAI. 2024.
\newblock {GPT-4 Technical Report}.
\newblock \emph{arXiv}.

\bibitem[{Pavlick and Kwiatkowski(2019)}]{pavlick2019inherent}
Ellie Pavlick and Tom Kwiatkowski. 2019.
\newblock \href {https://doi.org/10.1162/tacl_a_00293} {{Inherent Disagreements in Human Textual Inferences}}.
\newblock \emph{Transactions of the Association for Computational Linguistics}, 7:677--694.

\bibitem[{Ramprasad et~al.(2024)Ramprasad, Ferracane, and Lipton}]{ramprasad-etal-2024-analyzing}
Sanjana Ramprasad, Elisa Ferracane, and Zachary Lipton. 2024.
\newblock \href {https://doi.org/10.18653/v1/2024.acl-long.677} {Analyzing {LLM} behavior in dialogue summarization: Unveiling circumstantial hallucination trends}.
\newblock In \emph{Proceedings of the 62nd Annual Meeting of the Association for Computational Linguistics (Volume 1: Long Papers)}, pages 12549--12561, Bangkok, Thailand. Association for Computational Linguistics.

\bibitem[{Subbiah et~al.(2024{\natexlab{a}})Subbiah, Ladhak, Mishra, Adams, Chilton, and McKeown}]{subbiah2024storysumm}
Melanie Subbiah, Faisal Ladhak, Akankshya Mishra, Griffin~Thomas Adams, Lydia Chilton, and Kathleen McKeown. 2024{\natexlab{a}}.
\newblock \href {https://doi.org/10.18653/v1/2024.emnlp-main.557} {{{STORYSUMM}: Evaluating Faithfulness in Story Summarization}}.
\newblock In \emph{Proceedings of the 2024 Conference on Empirical Methods in Natural Language Processing}, pages 9988--10005, Miami, Florida, USA. Association for Computational Linguistics.

\bibitem[{Subbiah et~al.(2024{\natexlab{b}})Subbiah, Zhang, Chilton, and McKeown}]{subbiah2024reading}
Melanie Subbiah, Sean Zhang, Lydia~B. Chilton, and Kathleen McKeown. 2024{\natexlab{b}}.
\newblock \href {https://doi.org/10.1162/tacl_a_00702} {{Reading Subtext: Evaluating Large Language Models on Short Story Summarization with Writers}}.
\newblock \emph{Transactions of the Association for Computational Linguistics}, 12:1290--1310.

\bibitem[{Tang et~al.(2024)Tang, Laban, and Durrett}]{tang-etal-2024-minicheck}
Liyan Tang, Philippe Laban, and Greg Durrett. 2024.
\newblock \href {https://doi.org/10.18653/v1/2024.emnlp-main.499} {{{M}ini{C}heck: Efficient Fact-Checking of {LLM}s on Grounding Documents}}.
\newblock In \emph{Proceedings of the 2024 Conference on Empirical Methods in Natural Language Processing}, pages 8818--8847, Miami, Florida, USA. Association for Computational Linguistics.

\bibitem[{Uma et~al.(2022)Uma, Fornaciari, Hovy, Paun, Plank, and Poesio}]{uma2021learning}
Alexandra~N. Uma, Tommaso Fornaciari, Dirk Hovy, Silviu Paun, Barbara Plank, and Massimo Poesio. 2022.
\newblock \href {https://doi.org/10.1613/jair.1.12752} {{Learning from Disagreement: A Survey}}.
\newblock \emph{J. Artif. Int. Res.}, 72:1385–1470.

\bibitem[{Wadhwa et~al.(2024)Wadhwa, Zhao, Li, and Durrett}]{wadhwa-etal-2024-learning-refine}
Manya Wadhwa, Xinyu Zhao, Junyi~Jessy Li, and Greg Durrett. 2024.
\newblock \href {https://doi.org/10.18653/v1/2024.findings-emnlp.716} {{Learning to Refine with Fine-Grained Natural Language Feedback}}.
\newblock In \emph{Findings of the Association for Computational Linguistics: EMNLP 2024}, pages 12281--12308, Miami, Florida, USA. Association for Computational Linguistics.

\bibitem[{Wang et~al.(2024)Wang, Li, Shao, Xu, Dai, Li, Chen, Wu, and Sui}]{wang2023math}
Peiyi Wang, Lei Li, Zhihong Shao, Runxin Xu, Damai Dai, Yifei Li, Deli Chen, Yu~Wu, and Zhifang Sui. 2024.
\newblock \href {https://doi.org/10.18653/v1/2024.acl-long.510} {{Math-Shepherd: Verify and Reinforce {LLM}s Step-by-step without Human Annotations}}.
\newblock In \emph{Proceedings of the 62nd Annual Meeting of the Association for Computational Linguistics (Volume 1: Long Papers)}, pages 9426--9439, Bangkok, Thailand. Association for Computational Linguistics.

\bibitem[{Wang et~al.(2022)Wang, Wei, Schuurmans, Le, Chi, Narang, Chowdhery, and Zhou}]{wang2022self}
Xuezhi Wang, Jason Wei, Dale Schuurmans, Quoc Le, Ed~Chi, Sharan Narang, Aakanksha Chowdhery, and Denny Zhou. 2022.
\newblock {Self-Consistency Improves Chain of Thought Reasoning in Language Models}.
\newblock \emph{International Conference on Learning Representations (ICLR)}.

\bibitem[{Wei et~al.(2022)Wei, Wang, Schuurmans, Bosma, ichter, Xia, Chi, Le, and Zhou}]{wei2022chain}
Jason Wei, Xuezhi Wang, Dale Schuurmans, Maarten Bosma, brian ichter, Fei Xia, Ed~Chi, Quoc~V Le, and Denny Zhou. 2022.
\newblock \href {https://proceedings.neurips.cc/paper_files/paper/2022/file/9d5609613524ecf4f15af0f7b31abca4-Paper-Conference.pdf} {{Chain-of-Thought Prompting Elicits Reasoning in Large Language Models}}.
\newblock In \emph{Advances in Neural Information Processing Systems}, volume~35, pages 24824--24837. Curran Associates, Inc.

\bibitem[{Xu et~al.(2024{\natexlab{a}})Xu, Deutsch, Finkelstein, Juraska, Zhang, Liu, Wang, Li, and Freitag}]{xu-etal-2024-llmrefine}
Wenda Xu, Daniel Deutsch, Mara Finkelstein, Juraj Juraska, Biao Zhang, Zhongtao Liu, William~Yang Wang, Lei Li, and Markus Freitag. 2024{\natexlab{a}}.
\newblock \href {https://doi.org/10.18653/v1/2024.findings-naacl.92} {{{LLMR}efine: Pinpointing and Refining Large Language Models via Fine-Grained Actionable Feedback}}.
\newblock In \emph{Findings of the Association for Computational Linguistics: NAACL 2024}, pages 1429--1445, Mexico City, Mexico. Association for Computational Linguistics.

\bibitem[{Xu et~al.(2024{\natexlab{b}})Xu, Liu, Liu, Hou, Li, Zhang, Wang, Zeng, Du, Wenyi, Tang, and Dong}]{xu-etal-2024-chatglm}
Yifan Xu, Xiao Liu, Xinghan Liu, Zhenyu Hou, Yueyan Li, Xiaohan Zhang, Zihan Wang, Aohan Zeng, Zhengxiao Du, Zhao Wenyi, Jie Tang, and Yuxiao Dong. 2024{\natexlab{b}}.
\newblock \href {https://doi.org/10.18653/v1/2024.findings-emnlp.569} {{{C}hat{GLM}-Math: Improving Math Problem-Solving in Large Language Models with a Self-Critique Pipeline}}.
\newblock In \emph{Findings of the Association for Computational Linguistics: EMNLP 2024}, pages 9733--9760, Miami, Florida, USA. Association for Computational Linguistics.

\bibitem[{Yao et~al.(2023)Yao, Schloss, and Selvaraj}]{yao2023improving}
Zonghai Yao, Benjamin Schloss, and Sai Selvaraj. 2023.
\newblock \href {https://doi.org/10.18653/v1/2023.emnlp-main.158} {{Improving Summarization with Human Edits}}.
\newblock In \emph{Proceedings of the 2023 Conference on Empirical Methods in Natural Language Processing}, pages 2604--2620, Singapore. Association for Computational Linguistics.

\bibitem[{Ye et~al.(2023)Ye, Jo, Kim, Kim, Hwang, and Seo}]{selfee2023}
Seonghyeon Ye, Yongrae Jo, Doyoung Kim, Sungdong Kim, Hyeonbin Hwang, and Minjoon Seo. 2023.
\newblock \href {https://kaistai.github.io/SelFee/} {{SelFee: Iterative Self-Revising LLM Empowered by Self-Feedback Generation}}.
\newblock Blog post.

\bibitem[{Zelikman et~al.(2022)Zelikman, Wu, Mu, and Goodman}]{zelikman2022star}
Eric Zelikman, Yuhuai Wu, Jesse Mu, and Noah Goodman. 2022.
\newblock \href {https://proceedings.neurips.cc/paper_files/paper/2022/file/639a9a172c044fbb64175b5fad42e9a5-Paper-Conference.pdf} {{STaR: Bootstrapping Reasoning With Reasoning}}.
\newblock In \emph{Advances in Neural Information Processing Systems}, volume~35, pages 15476--15488. Curran Associates, Inc.

\bibitem[{Zhang et~al.(2023)Zhang, Liu, and Zhang}]{zhang-etal-2023-summit}
Haopeng Zhang, Xiao Liu, and Jiawei Zhang. 2023.
\newblock \href {https://doi.org/10.18653/v1/2023.findings-emnlp.714} {{{S}umm{I}t: Iterative Text Summarization via {C}hat{GPT}}}.
\newblock In \emph{Findings of the Association for Computational Linguistics: EMNLP 2023}, pages 10644--10657, Singapore. Association for Computational Linguistics.

\end{thebibliography}

\appendix
\section{Appendix}
\subsection{Synthetic data generation prompts}
The following prompts are filled with a story and summary or story and claim pair in the \%s fields.

\label{sec:appsynthprompts}
\begin{figure*}
\begin{tcolorbox}[
    colback=white,    
    colframe=myblue,  
    coltitle=white,   
    fonttitle=\bfseries, 
    title=Subjective $\rightarrow$ Objective Prompts,
    fontupper=\ttfamily
]
\small
\begin{Verbatim}[breaklines=true,breaksymbolleft={},breaksymbolright={},tabsize=0]
TYPE 1: Use the provided story to rewrite the provided claim to remove any ambiguous wording from the claim which may require or demonstrate some interpretation. If the claim can't be rewritten, give the original claim. Put the rewritten claim between <sentence> tags.\\Story: %s\\Claim: %s

TYPE 2: Use the provided story to rewrite the provided claim to remove any minor assumptions that the claim makes. If the claim can't be rewritten, given the original claim. Put the rewritten claim between <sentence> tags.\\Story: %s\\Claim: %s

TYPE 3: Use the provided story to rewrite the provided claim to not skip causal details or contain vague phrases that skip things. The provided claim is one of many summary claims, and must fit into the context of the summary when rewritten. If the claim can't be rewritten, give the original claim. Put the rewritten claim between <sentence> tags.\\Story: %s\\Summary: %s\\Rewrite only Line %s from the summary.

TYPE 4: Use the provided story to rewrite the provided claim to not specify any implied or ambiguous interpretations of the story as an explicit occurrence. If the claim can't be rewritten, given the original claim. Put the rewritten claim between <sentence> tags.\\Story: %s\\Claim: %s
\end{Verbatim}
\end{tcolorbox}
\end{figure*}

\begin{figure*}
\begin{tcolorbox}[
    colback=white,    
    colframe=myblue,  
    coltitle=white,   
    fonttitle=\bfseries, 
    title=Objective $\rightarrow$ Subjective Prompts,
    fontupper=\ttfamily
]
\small
\begin{Verbatim}[breaklines=true,breaksymbolleft={},breaksymbolright={},tabsize=0]
TYPE 1: Swap the wording in the claim in one or two places so it requires or demonstrates some interpretation of the story. The claim should become difficult to evaluate with respect to the story. You must rewrite the claim in some way. Put the rewritten claim between <sentence> tags.\\Story: %s\\Claim: %s

TYPE 2: Use the provided story to add a minor detail to the claim that isn't explicitly stated in the story. This detail must be a reasonable assumption to make from the story. The claim should become difficult to evaluate with respect to the story. You must rewrite the claim in some way. Put the rewritten claim between <sentence> tags.\\Story: %s\\Claim: %s

TYPE 3: Make the provided claim more vague about why things are happening by removing important causal details. The provided claim is one of many summary claims and must fit into the context of the summary when rewritten. The rewritten claim should become difficult to evaluate with respect to the story, but should not be shorter in length than the original claim. You must rewrite the claim in some way. Put the rewritten claim between <sentence> tags.\\Story: %s\\Summary: %s\\Rewrite only Line %s from the summary.

TYPE 4: Rewrite the claim to include some interpretation of what characters are thinking or feeling or what is happening in the story. State this definitively, rather than just as a possibility. The provided claim is one of many summary claims and must fit into the context of the summary when rewritten. The rewritten claim should become difficult to evaluate with respect to the story. You must rewrite the claim in some way. Put the rewritten claim between <sentence> tags.\\Story: %s\\Claim: %s\\Rewrite only Line %s from the summary.
\end{Verbatim}
\end{tcolorbox}
\end{figure*}

\subsection{Baseline and rewrite prompts}
The following prompts are filled with a story, summary, and claim triple in the \%s fields.
\label{sec:appbaselineprompts}
\begin{figure*}
\begin{tcolorbox}[
    colback=white,    
    colframe=myblue,  
    coltitle=white,   
    fonttitle=\bfseries, 
    title=Zero-shot prompt used as a baseline,
    fontupper=\ttfamily
]
\small
\begin{Verbatim}[breaklines=true,breaksymbolleft={},breaksymbolright={},tabsize=0]
SYSTEM: You are an expert summary evaluator, and you will be asked to evaluate claims in summaries of short stories. You will first be presented with the story and then the summary. You need to determine whether all of the information in the summary can be objectively evaluated for accuracy against the story or if there are claims that are subjective to evaluate. An objective claim may be accurate or inaccurate but it should be clearly right or wrong. A subjective claim introduces vague language, interpretation, or confusing wording such that different people might interpret it in different ways.
USER: Story:
%s

Summary:
%s

Consider the following claim in the summary: %s
Is this claim objective to evaluate? You should answer Yes or No. Place your answer between <answer></answer> tags.
\end{Verbatim}
\end{tcolorbox}
\end{figure*}

\begin{figure*}
\begin{tcolorbox}[
    colback=white,    
    colframe=myblue,  
    coltitle=white,   
    fonttitle=\bfseries, 
    title=Few-shot prompt used as a baseline,
    fontupper=\ttfamily
]
\small
\begin{Verbatim}[breaklines=true,breaksymbolleft={},breaksymbolright={},tabsize=0]
SYSTEM: You are an expert summary evaluator, and you will be asked to evaluate claims in summaries of short stories. You will first be presented with the story and then the summary. You need to determine whether all of the information in the summary can be objectively evaluated for accuracy against the story or if there are claims that are subjective to evaluate. An objective claim may be accurate or inaccurate but it should be clearly right or wrong. A subjective claim introduces vague language, interpretation, or confusing wording such that different people might interpret it in different ways.
USER: Story:
Shelly and her dog were running down the street one afternoon when they came across an injured squirrel. Shelly stopped to help the squirrel and Shelly's dog almost ate it. Shelly managed to tuck it into her pocket and bring it home. Later she brought it to a vet and got some recommendations on how to nurse it back to health. Within a couple weeks she released the squirrel back into the wild.

Summary: The main character, Shelly, and her dog find an injured squirrel while out running. The dog's prey drive is activated around the squirrel. Shelly tucks the little fluffy squirrel into her pocket to bring home. She figures out how to nurse it back to health. She eventually lets the squirrel go again to live a healthy life.

Consider the following claim in the summary: The main character, Shelly, and her dog find an injured squirrel while out running.
Is this claim objective to evaluate? You should answer Yes or No. Place your answer between <answer></answer> tags.
<answer>Yes</answer>

Consider the following claim in the summary: The dog's prey drive is activated around the squirrel.
Is this claim objective to evaluate? You should answer Yes or No. Place your answer between <answer></answer> tags.
<answer>No</answer>

Consider the following claim in the summary: Shelly tucks the little fluffy squirrel into her pocket to bring home.
Is this claim objective to evaluate? You should answer Yes or No. Place your answer between <answer></answer> tags.
<answer>No</answer>

Consider the following claim in the summary: She figures out how to nurse it back to health.
Is this claim objective to evaluate? You should answer Yes or No. Place your answer between <answer></answer> tags.
<answer>No</answer>

Consider the following claim in the summary: She eventually lets the squirrel go again to live a healthy life.
Is this claim objective to evaluate? You should answer Yes or No. Place your answer between <answer></answer> tags.
<answer>No</answer>

Story:
%s

Summary:
%s

Consider the following claim in the summary: %s
Is this claim objective to evaluate? You should answer Yes or No. Place your answer between <answer></answer> tags.
\end{Verbatim}
\end{tcolorbox}
\end{figure*}
\begin{figure*}
\begin{tcolorbox}[
    colback=white,    
    colframe=myblue,  
    coltitle=white,   
    fonttitle=\bfseries, 
    title=Self-consistency prompt used as a baseline,
    fontupper=\ttfamily
]
\small
\begin{Verbatim}[breaklines=true,breaksymbolleft={},breaksymbolright={},tabsize=0]
SYSTEM: You are an expert summary evaluator, and you will be asked to evaluate claims in summaries of short stories. You will first be presented with the story and then the summary. You need to determine whether all of the information in the summary is consistent with the information in the story. The details described in a consistent summary should not misrepresent details of the story or make things up. 
USER: Story:
%s

Summary:
%s

Consider the following claim in the summary: %s
Is all of the information in this claim consistent with the story? First reason about the question before answering Yes or No. Your output should be in the following format:

Reasoning: Your reasoning about the answer to the question.

<answer>Your answer to the question (Yes or No)</answer>
\end{Verbatim}
\end{tcolorbox}
\end{figure*}

\begin{figure*}
\begin{tcolorbox}[
    colback=white,    
    colframe=myblue,  
    coltitle=white,   
    fonttitle=\bfseries, 
    title=Rewrite prompt used for subjectivity-focused rewrites,
    fontupper=\ttfamily
]
\small
\begin{Verbatim}[breaklines=true,breaksymbolleft={},breaksymbolright={},tabsize=0]
SYSTEM: You are an expert summary writer. You write and correct summaries so that they are precise, clear and accurate representations of the story.
USER: Story:
%s

Summary:
%s

Rewrite any elements of the following sentence from the summary that might be subjective. You should make minimal edits to fix the sentence. If the sentence is objective as written or is just interpretation of the story, restate the original sentence. Give your final sentence in <answer></answer> tags.

Sentence: %s
\end{Verbatim}
\end{tcolorbox}
\end{figure*}

\begin{figure*}
\begin{tcolorbox}[
    colback=white,    
    colframe=myblue,  
    coltitle=white,   
    fonttitle=\bfseries, 
    title=Rewrite prompt used for unfaithfulness-focused rewrites,
    fontupper=\ttfamily
]
\small
\begin{Verbatim}[breaklines=true,breaksymbolleft={},breaksymbolright={},tabsize=0]
SYSTEM: You are an expert summary writer. You write and correct summaries so that they are precise, clear and accurate representations of the story.
USER: Story:
%s

Summary:
%s

Rewrite any elements of the following sentence from the summary that are inconsistent with the story. You should make minimal edits to fix the sentence. If the sentence is accurate as written or is just interpretation of the story, restate the original sentence. Give your final sentence in <answer></answer> tags.

Sentence: %s
\end{Verbatim}
\end{tcolorbox}
\end{figure*}

\begin{figure*}
\begin{tcolorbox}[
    colback=white,    
    colframe=myblue,  
    coltitle=white,   
    fonttitle=\bfseries, 
    title=Rewrite prompt targeting both subjectivity and faithfulness (ARM),
    fonttitle=\small
]
\small
\begin{Verbatim}[breaklines=true,breaksymbolleft={},breaksymbolright={},tabsize=0]
SYSTEM: You are an expert summary writer. You write and correct summaries so that they are precise, clear and accurate representations of the story.

USER: Read the story and summary carefully, then decide whether the specified summary sentence should be rewritten.\\
A summary sentence should be rewritten according to the following principles:
1.) If the sentence is inconsistent with the story, it should be rewritten.
2.) If the sentence contains subjective interpretation or ambiguous wording, it should be rewritten. In particular, rewrite cases of:
    - assuming a minor detail that is reasonable but not explicitly stated in the story
    - skipping important causal details
    - making explicit conclusions which are left ambiguous in the story
    - using words or phrases that can be interpreted differently from the story wording
3.) When rewriting a sentence, any edits should be minimal to fix the problem.
4.) If the sentence is just commentary on the story, then it should not be rewritten.
5.) If the sentence is accurate and clear, it should not be rewritten.\\
Story:\\%s\\\\Summary:\\%s\\
Rewrite the following summary sentence, placing your rewrite between <answer></answer> tags. If the sentence does not need to be rewritten, simply repeat the original wording between <answer></answer> tags.\\
Sentence: %s
\end{Verbatim}
\end{tcolorbox}
\end{figure*}







\subsection{Explanation parsing prompt}
\label{sec:appexpprompt}
\begin{tcolorbox}[
    colback=white,    
    colframe=myblue,  
    coltitle=white,   
    fonttitle=\bfseries, 
    title=Prompt for parsing an explanation into individual points,
    fontupper=\ttfamily
]
\small
\begin{Verbatim}[breaklines=true,breaksymbolleft={},breaksymbolright={},tabsize=0]
Summarize the key reasons described in this explanation for why the summary sentence needs to be rewritten. Group together reasoning about the same detail. Place each reason between <item></item> tags.
\end{Verbatim}
\end{tcolorbox}

\subsection{Annotation format validation}
\label{sec:appannovalid}
We check for confounding factors in the task format for the human annotations of rewrites in several ways. First, we randomize whether the rewrite is shown to the annotator in the summary or as the ``alternate''. We find there is not a significant difference ($p$: .08) in which claim is preferred based on which position it is shown to annotators in (57.1\% for in-summary vs. 42.9\% for alternate). These numbers indicate annotators are not biased to prefer whichever claim is shown to them as the ``original'' vs. ``alternate''. 

Additionally, we include three decoy explanations that do not make sense and annotators should reject to check that annotators are not just convinced by any explanation. The decoy explanations are marked as WRONG in 5/6 instances indicating that annotators are not just convinced by any explanation. We remove results on these decoy explanations from the results discussed in the paper.

\subsection{Human annotation interfaces}
\begin{figure*}[h!]
\centering
    \includegraphics[width=\linewidth]{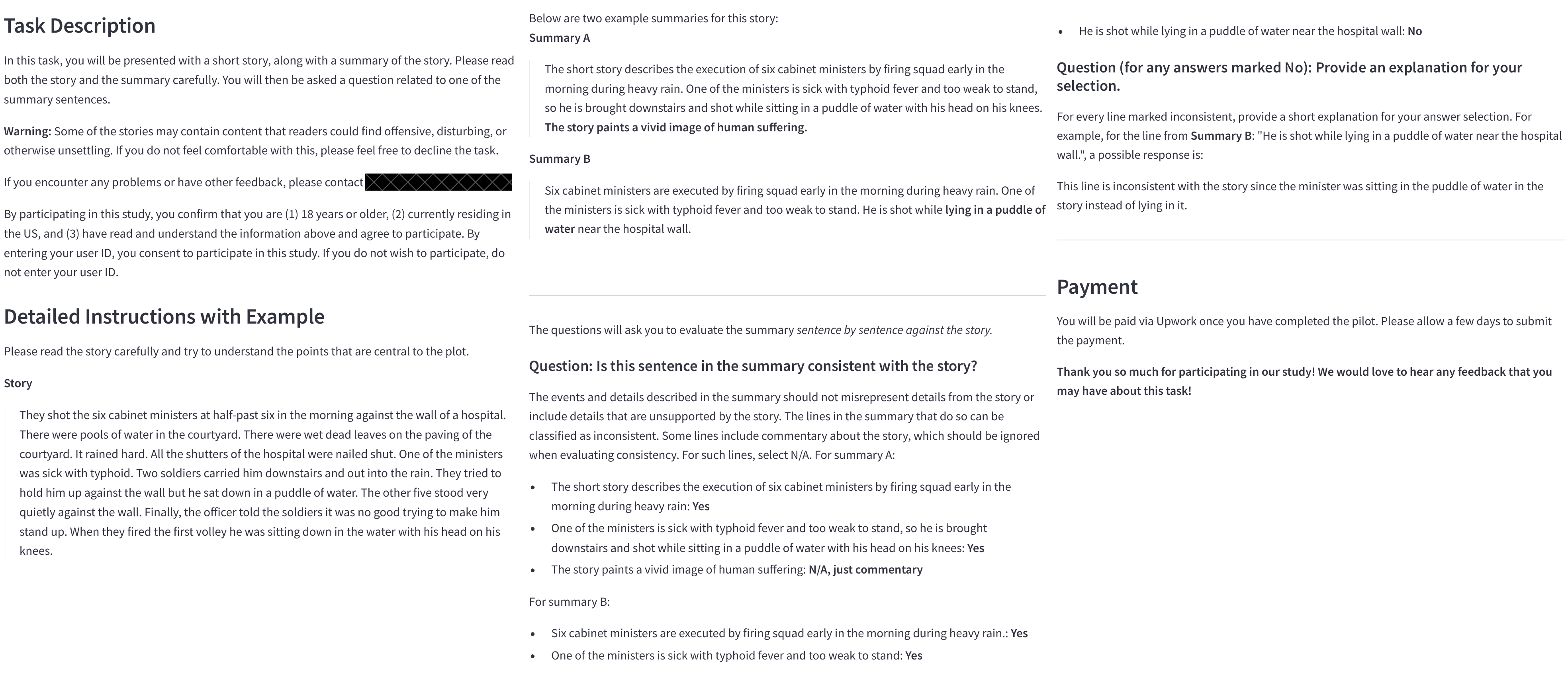}
\caption{A variant of these instructions are used for each of the human annotation tasks on Upwork.}
    \label{fig:appinstructions}
\end{figure*}

\begin{figure*}[h!]
\centering
    \includegraphics[width=\linewidth]{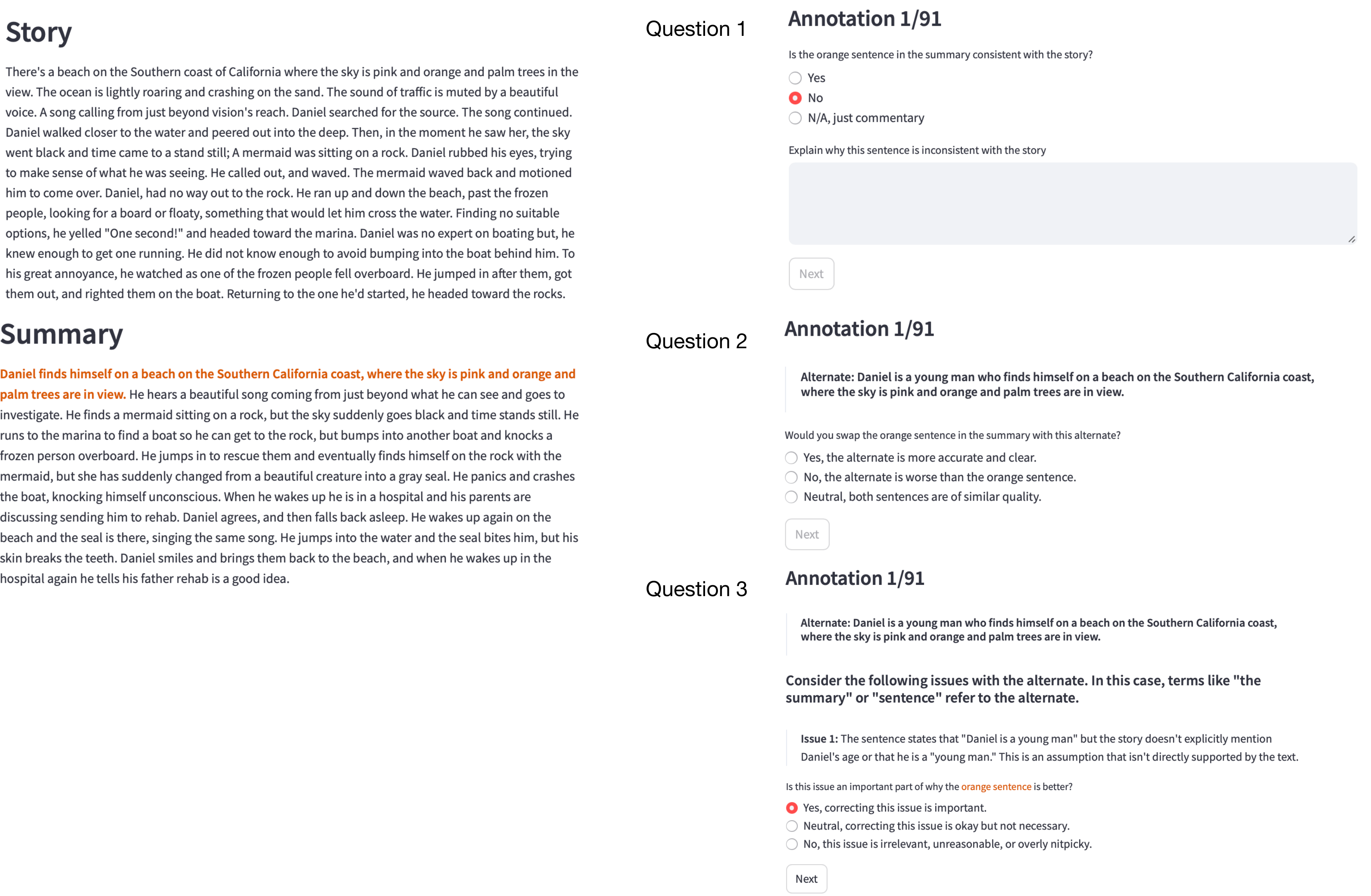}
\caption{The task format for human annotations on Upwork. Human studies which only look at annotator agreement on faithfulness labels only ask Question 1. The studies aligning rewrites with annotator judgments additionally ask questions 2 and 3.}
    \label{fig:apptasks}
\end{figure*}

Screenshots from the Upwork annotation interfaces are shown in Figures \ref{fig:appinstructions} and \ref{fig:apptasks}.

\subsection{Additional ambiguity analysis details}
\label{sec:appambiganalysis}
For the inductive thematic analysis, the two authors involved write an explanation for the ambiguities involved in each subjective claim. They then discuss these explanations and arrive at four types of ambiguities. They go back and code each subjective claim with one of these types and discuss and adjudicate any disagreements. We observed agreement between the authors on 89\% of the sentences (0.53 Cohen's Kappa) prior to adjudication.

They additionally label a small number of claims with a type 5 not discussed in the main text. 
Type 5 indicates the story is too confusing to determine its intent but the summary sentence itself is written clearly. We do not include type 5 claims in the analysis as they result from unintentional ambiguity introduced by the story writer which is beyond the scope of this paper.

While the writer of a story may intentionally use ambiguity in their story which leads to subjective viewpoints on its meaning, a summary writer should not intentionally introduce ambiguity. Types 1-3 capture types of ambiguities in summary claims that are unintentional on the part of the summary writer. 
Type 4 deals with intentional ambiguity by the author. We observe that type 4 ambiguities may be okay in summaries that should interpret the story as well as summarizing it, while types 1-3 are always undesirable as they obscure meaning. 

In Figure \ref{fig:appbarcharts}, we show the recall of Claude and GPT-4 rewrite metrics averaged across the three different rewriting prompts for each ambiguity type. We do not observe a significant difference in recall for the different types.

\begin{figure}[]
\centering \small
\begin{tikzpicture}
    \begin{axis}[
        ybar,
        bar width=5pt,
        width=\columnwidth,
        height=4cm,
        symbolic x coords={1, 2, 3, 4},
        xtick=data,
        ytick={0,.25, .5, .75, 1},
        ymin=0,
        ymax=1,
        xlabel={Ambiguity type },
        ylabel={Recall},
        ymajorgrids=true,
        enlarge x limits=0.3,
        legend pos = south east,
        major tick length=0pt,
        legend style={nodes={scale=0.8, transform shape}}
    ]
        \addplot coordinates {(1, 0.75) (2, 0.88) (3, 0.84) (4, 0.82)};
        \addplot coordinates {(1, 0.76) (2, 0.76) (3, 0.82) (4, 0.76)};
        \legend{GPT-4, Claude-3.5}
    \end{axis}
\end{tikzpicture}
\caption{We show the recall of GPT-4 rewrite metrics vs. Claude-3.5 rewrite metrics at detecting different ambiguity types. For each model, we average the results across the three rewrite prompts tested in Table \ref{tab:rewriteresults}.}
\label{fig:appbarcharts}
\end{figure}

\end{document}